\documentclass[runningheads]{llncs}

\usepackage{eccv}
\usepackage{eccvabbrv}
\usepackage{graphicx}
\usepackage{booktabs}
\usepackage{algorithm}
\usepackage{algorithmicx}
\usepackage{algpseudocode}
\usepackage[utf8]{inputenc}
\usepackage[T1]{fontenc}
\usepackage{amsfonts}
\usepackage{nicefrac}
\usepackage{microtype}
\usepackage{xcolor}
\usepackage{comment}
\usepackage{arydshln}
\usepackage{enumitem}
\usepackage[normalem]{ulem}
\usepackage{colortbl}
\usepackage{nicematrix}
\usepackage{lipsum}
\usepackage{cancel}
\usepackage{tabularx}
\usepackage{multicol}
\usepackage{multirow}
\usepackage{amssymb}
\usepackage{bm}
\usepackage{subcaption}
\usepackage{wrapfig}
\usepackage{indentfirst}
\usepackage[accsupp]{axessibility}
\usepackage{hyperref}
\usepackage{orcidlink}

\makeatletter
\renewcommand*{\@fnsymbol}[1]{\ensuremath{\ifcase#1\or \dagger \or \ddagger \or \mathsection \else\@ctrerr\fi}}
\makeatother

\begin{document}

\title{DARE to Mitigate Hallucination:\\Dual-path Auto-Regressive-aware Editing}

\titlerunning{Dual-path Auto-Regressive-aware Editing}

\author{Jae-Ho Lee\orcidlink{0009-0003-0960-824X} \and
Jeong-Eun Lee\orcidlink{0009-0007-8655-6438} \and
Gyeong-Moon Park\thanks{Corresponding Author.}\orcidlink{0000-0003-4011-9981}}

\authorrunning{J.-H. Lee et al.}

\institute{Korea University, Seoul, Republic of Korea\\
\email{\{jaeho-lee, esilver, gm-park\}@korea.ac.kr}}

\maketitle
\begin{abstract}
Large vision-language models (LVLMs) have recently achieved remarkable progress across multimodal tasks, yet object hallucination remains a persistent challenge where models generate descriptions inconsistent with the visual input. Recent work mitigates hallucinations through training-free representation editing, typically by constructing hallucination-related directions from teacher-forcing (TF) contrasts between hallucinated and truthful responses. However, LVLMs operate through autoregressive (AR) decoding during generation, raising the question of whether TF-based analysis fully reflects the generation dynamics that lead to hallucinated outputs. In this paper, we analyze the relationship between TF-based editing and AR generation behavior and find that TF-based editing alone may be insufficient to capture both decoding dynamics and multimodal interactions associated with hallucinations. To address this limitation, we propose \textbf{DARE (Dual-path Auto-Regressive-aware Editing)}, a hybrid hallucination editing framework that integrates two complementary contrast pathways: textual contrasts and image contrasts, together with autoregressive-aware representation signals. Specifically, DARE constructs hallucination editing directions from (1) TF-based textual contrasts, (2) AR-aware representation transitions during decoding, and (3) controlled visual differences between paired images. Extensive experiments on multiple LVLM hallucination benchmarks demonstrate that DARE consistently reduces object hallucinations while preserving multimodal perception capability and inference efficiency. Our implementation code is available at \url{https://github.com/KU-VGI/DARE}.
\end{abstract}
\section{Introduction}
\label{sec:introduction}

\begin{figure}[!t]
    \centering
    \includegraphics[width=\linewidth]{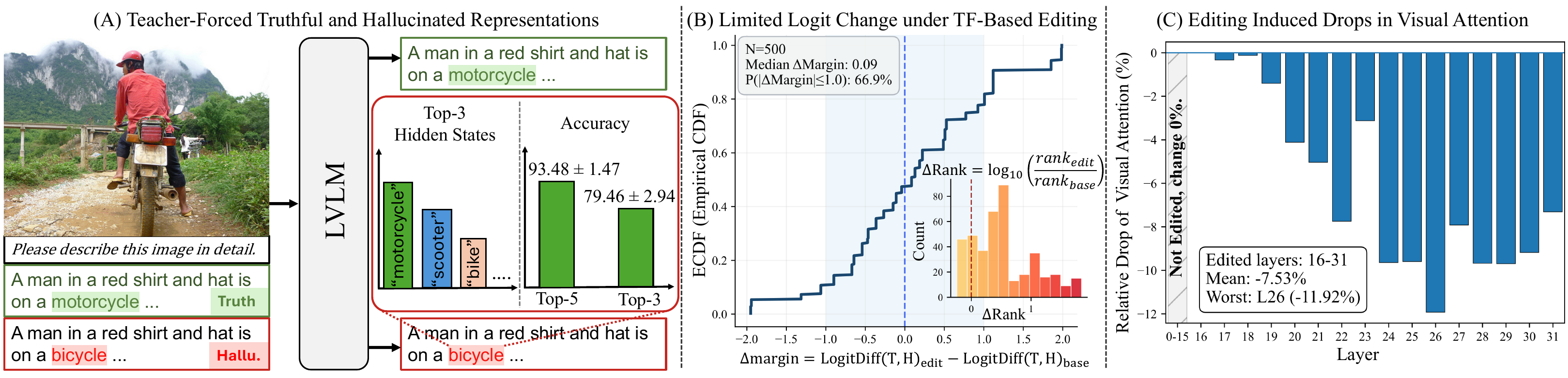}
    \caption{Limitations of teacher-forcing (TF)-based hallucination editing in LVLMs. 
    (A) Under TF evaluation, truthful tokens frequently remain among the top-ranked candidates even when evaluated on hallucinated captions under TF. 
    (B) TF-based editing modifies token distributions but results in limited changes in the logit margin between truthful and hallucinated tokens, indicating that generation decisions often remain unchanged. The inset histogram further shows substantial shifts in token rankings. 
    (C) Editing directions derived from textual contrasts are accompanied by reduced attention to visual tokens across layers. 
    These observations highlight limitations of TF-based hallucination editing and motivate approaches that explicitly consider generation dynamics and multimodal representations.}
    \label{fig:motivation}
\end{figure}

Large Vision-Language Models (LVLMs)~\cite{liu2023improved, zhu2023minigpt4, ye2024mplug} have recently achieved remarkable progress across a wide range of multimodal tasks, including visual question answering, image captioning, and multimodal dialogue. Despite these advances, \textbf{Object Hallucination (OH)} \cite{rawte2023hallucination, liu2024mitigating, liu2023mitigating, li2023evaluating, kim2023exposing, jiang2024hallucination} remains a persistent challenge, where models generate textual content that is inconsistent with objects present in the visual input. Such hallucinations reduce the reliability of LVLMs and limit their deployment in real-world applications \cite{heo2026detecting,park2025sfuod,choe2025universal,choe2024open}. Consequently, a growing body of research~\cite{leng2024mitigating , huang2024opera, wan2025only, fang2025grounding, huo2025selfintrospective, yin2025clearsight, yang2025nullu} has focused on understanding the causes of hallucinations and developing techniques to mitigate them.

\noindent Among recent approaches, representation editing \cite{jiang2024interpreting, yang2025nullu, yang2025mitigating} has emerged as a practical strategy for hallucination mitigation. Instead of retraining \cite{yu2024rlhf, fu2025mitigating, zhao2023beyond}, these methods perform \textit{training-free weight editing} by identifying directions associated with hallucinated responses. In particular, several approaches \cite{yang2025nullu, duan2025truthprint} construct a hallucination-related subspace (hereafter HalluSpace) by contrasting truthful and hallucinated responses under teacher-forcing (TF). The model is then edited through null-space projection that removes hallucination-associated components from the representation space. This approach enables hallucination mitigation without additional training while preserving inference efficiency.

\noindent To better understand how TF-based editing operates, we first analyze token distributions under TF evaluation. Surprisingly, we observe that truthful tokens often remain highly ranked under TF evaluation, even when conditioning on hallucinated captions. As illustrated in Fig.~\ref{fig:motivation}(A), the ground-truth token frequently appears within the top-ranked candidates of hallucinated representations. In particular, we measure how often the truthful token appears among the top-$k$ candidates of the hidden representation, and find that it remains within the top-5 candidates with high probability. This observation indicates that the model often retains a strong internal preference toward the correct token even in hallucination cases. Therefore, hallucinated generations may arise not from the absence of correct knowledge in the representation, but from how relative token scores are compared and selected during auto-regressive generation.

\noindent Motivated by this observation, we investigate whether TF-based editing actually changes the token selection during auto-regressive generation. If TF-based editing effectively reduces hallucinations, it should shift the model’s preference from the hallucinated token to the truthful token at decoding time. To test this, we compute the logit margin between the truthful token and the hallucinated token before and after editing on MSCOCO~\cite{lin2014microsoft}. Fig.~\ref{fig:motivation}(B) shows the empirical cumulative distribution function (ECDF) of the margin difference between the edited and base models for hallucinated samples generated under auto-regressive decoding. We find that most margin differences are close to zero, meaning that TF-based editing only slightly changes the relative scores of the two tokens. Since autoregressive decoding selects the token with the highest logit, the generated output changes only if the truthful token overtakes the hallucinated token in rank. However, when the margin remains nearly unchanged, the hallucinated token often continues to have the higher score and is therefore still selected. This analysis suggests that although TF-based editing can modify hidden representations and slightly adjust token distributions, it often fails to change the final decoding decision. As a result, hallucinations may persist even when the truthful token is ranked highly in the model’s internal representation.

\noindent Beyond generation dynamics, we examine whether hallucination directions derived from \emph{text-only} contrasts can inadvertently affect multimodal interactions in LVLMs. Most prior methods \cite{jiang2024interpreting, yang2025nullu, yang2025mitigating, duan2025truthprint} construct editing directions by contrasting hallucinated and truthful textual responses, implicitly treating hallucination as primarily a linguistic difference. However, Fig.~\ref{fig:motivation} (A) shows that even for hallucinated evaluation, the model often retains strong evidence for the truthful token under teacher-forcing, suggesting that image-conditioned information remains present in the internal representation. Since such representations are already multimodal and entangled, text-only editing directions may correlate with components that also support visual-token interactions. Consistently, Fig.~\ref{fig:motivation} (C) shows that applying text-only editing is accompanied by reduced attention to visual tokens across layers, indicating a potential collateral effect on visual pathways when editing is defined solely from textual contrasts.

\noindent Based on these observations, we propose a hybrid editing framework coined \textbf{DARE (Dual-path Auto-Regressive-aware Editing)} designed to address both limitations. DARE is built on two key principles: \emph{dual-path contrasts} and \emph{autoregressive-aware editing}. First, DARE constructs hallucination editing directions through \textbf{two complementary contrast pathways}. The first pathway exploits \textbf{textual contrasts} between hallucinated and truthful responses under TF evaluation, capturing linguistic representation differences associated with hallucination. The second pathway utilizes \textbf{image contrasts} (via \textbf{Visual-Difference Editing}), where controlled visual differences between paired images induce representation changes grounded in visual evidence. DARE integrates textual and AR-aware directions into an \textbf{Augmented HalluSpace} and complements it with \textbf{Visual-Difference Editing} based on image contrasts. Empirically, DARE consistently reduces hallucination rates across multiple LVLM benchmarks while maintaining generation quality. Further analyses show improved stability of generation decisions and more consistent multimodal behaviors during autoregressive generation. Our contributions are summarized as follows:
\begin{itemize}
\item We provide a systematic analysis of TF-based hallucination editing methods and show that (i) truthful tokens often remain highly ranked under TF evaluation even when hallucinations occur, (ii) despite reshaping token distributions, TF-based editing frequently yields only limited changes in token-level decoding decisions under AR generation, and (iii) text-only editing directions can be accompanied by reduced attention to visual tokens across layers.
\item Based on these analyses, we propose \textbf{DARE (Dual-path Auto-Regressive-aware Editing)}, a training-free weight editing framework that integrates \textbf{dual-path contrasts}, \textbf{text contrasts} and \textbf{image contrasts} and further incorporates \textbf{AR-aware signals} derived from AR decoding trajectories to better target hallucination-inducing behaviors.
\item We conducted extensive experiments on multiple LVLM hallucination benchmarks, demonstrating that DARE consistently reduces object hallucinations while preserving multimodal perception capability and inference efficiency.
\end{itemize}

\section{Related Work}
\label{sec:relatedworks}
\noindent \textbf{Large Vision-Language Models.}
Large Vision-Language Models (LVLMs) have rapidly evolved following the success of Large Language Models (LLMs).  This progress, in turn, has led to significant improvements in multimodal tasks such as image captioning\cite{you2016image, vinyals2015show, anderson2018bottom} and visual question answering\cite{antol2015vqa, singh2019towards}. Specifically, to further enhance multimodal understanding, LLaVA\cite{liu2024visual} and LLaVA-1.5\cite{liu2023improved} integrate pretrained vision encoders with language decoders and employ instruction tuning\cite{peng2023instruction} to improve visual reasoning and comprehension. MiniGPT-4\cite{zhu2023minigpt4} utilizes a Q-former\cite{li2023blip} to efficiently fuse visual and textual features while reducing redundant visual tokens. mPLUG-Owl2\cite{ye2024mplug} adopts a parameter-efficient adaptation strategy that trains image-conditioned visual prompts instead of performing full model fine-tuning.
Beyond these representative architectures, more advanced line of LVLM models \cite{li2023blip, bai2023qwen, chen2023shikra, chen2024internvl, kim-etal-2026-open} have emerged.
\\
\\
\noindent \textbf{Mitigating Object Hallucination in LVLMs.} 
Object hallucination, which refers to the inconsistency between the objects in visual content and the generated text description\cite{rohrbach2018object}, has been a persistent challenge in LVLMs. 
Approaches for mitigating hallucinations can generally be categorized according to the stage at which they are applied.
The first line of research targets the training stage, where models are improved with newly optimized training objectives\cite{dai2023plausible, lyu2024alleviating, sarkar2025mitigating} or through the use of task-specifically constructed datasets\cite{peng2025mitigating, liu2024investigating} for OH mitigation. Although these approaches have demonstrated promising results, they require extensive computational resources and time.
Naturally, another line of research has emerged, that focuses on inference-stage techniques including advanced decoding–based strategies \cite{li2022contrastive, leng2024mitigating, chen2024halc, chuang2023dola, huang2024opera, wan2025only, huo2025selfintrospective} as well as attention intervention methods~\cite{liu2024paying, yin2025clearsight}. In addition, some recent works have attempted to mitigate hallucinations by editing model weights or steering the latent space. 
While latent steering methods \cite{liureducing,lee2025esc,seo2024generative,lee2026perturb} adjust latent features by adding precomputed stable directions at inference time, weight-editing approaches \cite{yang2025nullu,duan2025truthprint} project model weights onto the null space of hallucination-related components to explicitly attenuate them. Our work addresses hallucination by refining the weights of LVLMs, but in a way that more effectively preserves multimodal dependencies.
\\
\\
\noindent \textbf{Teacher-forcing and Auto-regressive Generation Gap.}
The discrepancy between teacher-forcing (TF) and auto-regressive (AR) generation has long been studied in language sequence modeling as the exposure bias problem \cite{bengio2015scheduled, ranzato2015sequence, schmidt2019generalization, wang2020exposure, arora2022exposure}. During TF, models are conditioned on ground-truth tokens, whereas during AR they rely on their own previously generated outputs. This mismatch can lead to cascading errors and unstable generation dynamics. While exposure bias has been extensively studied in language sequence modeling, its implications for hallucinated generation in LVLMs remain unclear. In particular, several hallucination editing methods derive editing directions by analyzing hidden representations under TF settings. However, hallucinated outputs arise during AR generation, where token selection depends on relative score comparisons among candidate tokens. This discrepancy suggests that editing directions derived from teacher-forced representations may not fully reflect the generation dynamics that influence token selection during generation. Our work revisits this gap and analyzes its implications for hallucination mitigation in LVLMs.
\section{Method}
\label{sec:method}
\begin{figure}[!t]
    \centering
    \includegraphics[width=\textwidth]{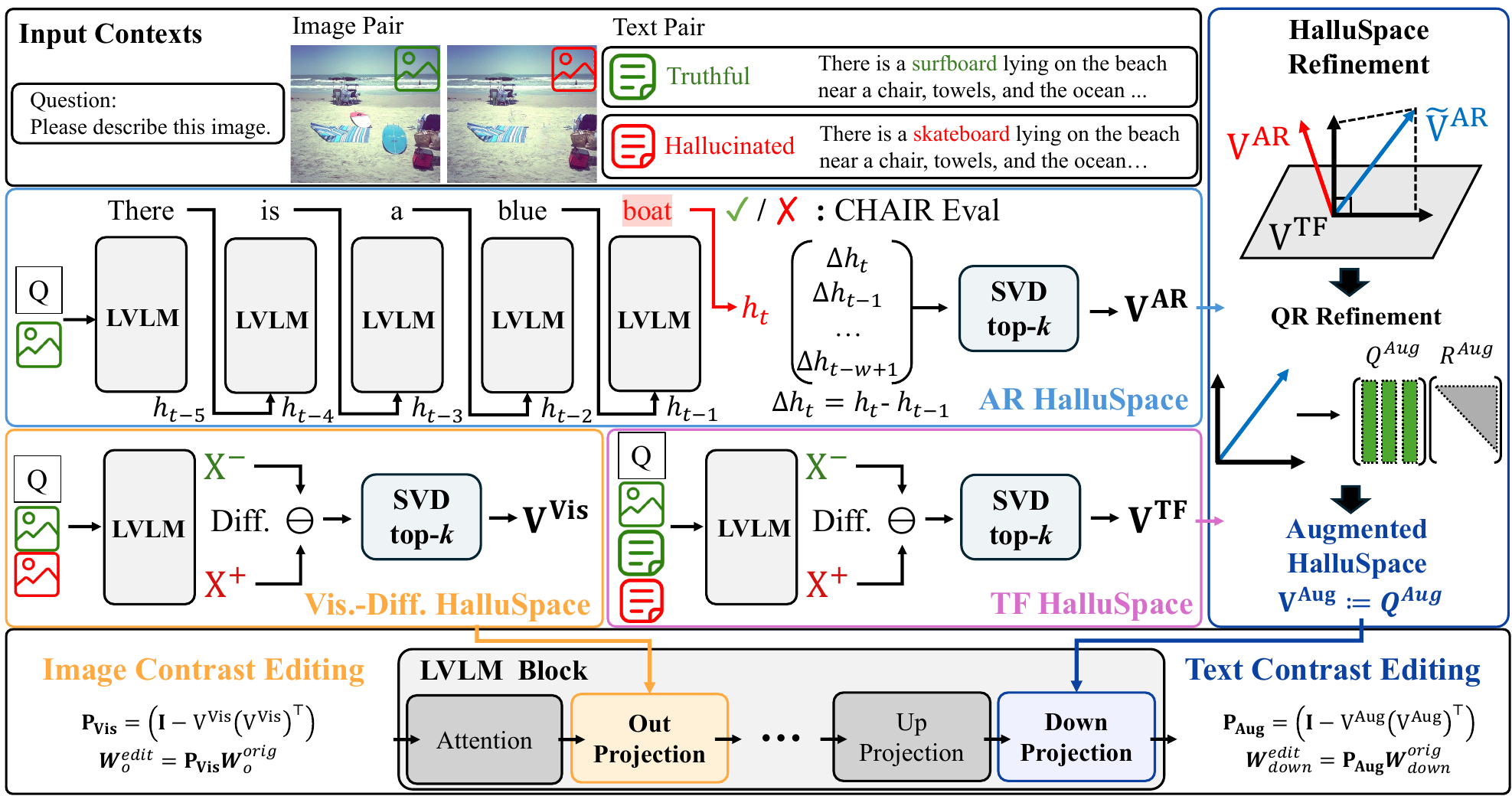}
    \caption{Architecture overview of DARE. DARE constructs AR-aware, TF-based, and visual-difference HalluSpaces, refines the TF and AR directions into an Augmented HalluSpace, and applies module-specific null-space editing to mitigate hallucination.}
    \label{fig:main}
\end{figure}

\noindent This section introduces \textbf{DARE (Dual-path Auto-Regressive-aware Editing)}, a hallucination mitigation method for LVLMs, as illustrated in Fig.~\ref{fig:main}. DARE constructs hallucination editing directions through two complementary contrast pathways together with autoregressive-aware signals. Specifically, DARE leverages (1) \textit{textual contrasts} derived from hallucinated and truthful responses under teacher-forcing, and (2) \textit{image contrasts} obtained from controlled visual differences. In addition, DARE incorporates \textit{Auto-Regressive-aware} signals that explicitly capture representation transitions occurring during decoding. Together, these components enable DARE to identify hallucination-related representation directions while preserving visually grounded multimodal interactions. 
\par\smallskip
\noindent We first review TF-based hallucination editing methods that construct hallucination subspaces from textual representation contrasts. We then introduce the \textit{AR-aware HalluSpace}, which derives hallucination directions from autoregressive generation trajectories. Finally, we present the full DARE framework that integrates textual contrasts, autoregressive dynamics, and visual contrasts.

\subsection{Preliminary: TF-based HalluSpace Editing}
\label{sec:preliminaries}

\noindent Consider a large vision-language model (LVLM) that generates a token sequence $y=(y_1,\dots,y_T)$ autoregressively conditioned on an image $I$ and a prompt $p$.
Object hallucination refers to cases where the generated text describes objects that do not exist in the input image. Recent editing approaches attempt to mitigate hallucinations by identifying latent representation directions associated with hallucinated outputs and removing them from the model parameters.
In particular, existing work~\cite{yang2025nullu} constructs a hallucination subspace by contrasting hallucinated and truthful responses under teacher-forcing evaluation. Formally, consider a dataset of paired responses,
\begin{equation}
\mathcal{D}=\{(x_i^{+},x_i^{-})\}_{i=1}^{N},
\end{equation}
where $x_i^{+}$ and $x_i^{-}$ denote hallucinated and truthful responses respectively.
Teacher-forcing is applied to obtain hidden representations at layer $\ell$.
Let $h_\ell(x)$ denote the pooled hidden representation of sequence $x$ at layer $\ell$.
Stacking representations across samples yields matrices $X_\ell^{+}, X_\ell^{-}\in\mathbb{R}^{N\times D}$. The hallucination contrast matrix is defined as
\begin{equation}
E_\ell = X_\ell^{+}-X_\ell^{-}.
\end{equation}

\noindent Applying singular value decomposition (SVD), $E_\ell = U_\ell\Sigma_\ell V_\ell^\top$, the top-$k$ right singular vectors form the TF-based hallucination subspace as
\begin{equation}
    V_\ell^{TF}=[v^{TF}_1,\dots,v^{TF}_{k_{TF}}]
    \in \mathbb{R}^{D\times k_{\mathrm{TF}}}.
\end{equation}
The model weights are then edited through null-space projection,
\begin{equation}
W_\ell^{edit}=(I-V_\ell^{TF}(V_\ell^{TF})^\top)W_\ell^{orig},
\end{equation}
which removes hallucination-related representation without additional training.

\noindent However, TF-based HalluSpace is derived entirely from teacher-forced representations, whereas hallucinated outputs are produced during autoregressive decoding, where next-token decisions depend on local competitions among candidate tokens. As discussed in Sec.~\ref{sec:introduction}, this mismatch motivates incorporating AR-aware signals that reflect decoding-time dynamics.

\subsection{AR-aware HalluSpace Construction}
\label{sec:ar_halluspace}

\noindent To better capture hallucination dynamics during generation, we construct hallucination directions directly from autoregressive decoding trajectories.
We use the LURE dataset~\cite{zhou2023analyzing} as prior work. Given an image $I$ and the question $q=\text{``Please describe this image in detail.''}$, the LVLM generates a caption autoregressively as $y=(y_1,\dots,y_T)$. Following the CHAIR~\cite{rohrbach2018object} evaluation protocol, hallucinated objects in the generated caption are detected.
Since CHAIR evaluates hallucination at the object level, we can identify the token position $t$ where the hallucinated object first appears.
Let $h_{\ell,j}\in\mathbb{R}^D$ denote the hidden representation at layer $\ell$ when generating token $y_j$.
Instead of contrasting teacher-forced responses, we analyze local representation transitions around the hallucination position. Specifically, we compute sequential differences $\Delta h_{\ell,j}=h_{\ell,j}-h_{\ell,j-1}$.
For each hallucination event at position $t$, we collect a short temporal window
$\{\Delta h_{\ell,t},\Delta h_{\ell,t-1},\dots,\Delta h_{\ell,t-w+1}\}$, which captures a short context of decoding-time state transitions leading into the hallucinated token. These vectors capture the representation dynamics immediately preceding hallucinated token generation. Collecting transitions from all hallucination samples forms the matrix
\begin{equation}
G_\ell =
[\Delta h_\ell^{(1)}, \Delta h_\ell^{(2)}, \dots, \Delta h_\ell^{(N_{\mathrm{AR}})}]^\top
\in \mathbb{R}^{N_{\mathrm{AR}}\times D},
\end{equation}
where $N_{\mathrm{AR}}$ denotes the total number of collected transitions.
Applying SVD to $G_\ell$ yields $G_\ell=U_\ell\Sigma_\ell V_\ell^\top$.
Since transition vectors are stacked row-wise in $G_\ell$, the right singular vectors in $V_\ell$ correspond to feature-space directions.
The top-$k_{\mathrm{AR}}$ right singular vectors define the AR-aware hallucination subspace as
\begin{equation}
V_\ell^{\mathrm{AR}}=[v^{\mathrm{AR}}_1,\dots,v^{\mathrm{AR}}_{k_{\mathrm{AR}}}]
\in \mathbb{R}^{D\times k_{\mathrm{AR}}}.
\end{equation}

\noindent Unlike TF-based HalluSpace, these directions capture autoregressive state transitions that occur immediately before hallucinated tokens are produced.

\subsection{Augmented HalluSpace Editing}
\label{sec:hybrid_editing}

\noindent The TF-based HalluSpace and the AR-aware HalluSpace capture different aspects of hallucinated generation. To leverage these complementary signals, we construct an \textit{Augmented HalluSpace} that integrates both TF-based and AR-based hallucination directions.
Let $V_\ell^{\mathrm{TF}}$ denote the TF HalluSpace obtained from teacher-forcing contrasts and $V_\ell^{\mathrm{AR}}$ denote the AR HalluSpace derived from autoregressive generation trajectories. Before combining the two sets of directions, we residualize the AR basis vectors with respect to the TF subspace.
This step ensures that AR directions capture information that is not already explained by TF-based components and provides a non-redundant basis for null-space editing.
Concretely, each AR vector $v^{\mathrm{AR}}_j$ is projected onto the orthogonal complement of the TF subspace as
\begin{equation}
\tilde v^{\mathrm{AR}}_j =
\left(I - V_\ell^{\mathrm{TF}}(V_\ell^{\mathrm{TF}})^\top\right)v^{\mathrm{AR}}_j .
\end{equation}
The resulting vectors are stacked to form the residualized AR basis $\tilde V_\ell^{\mathrm{AR}}$. We then concatenate the TF HalluSpace and the residualized AR directions and apply QR decomposition:
\begin{equation}
\left[V_\ell^{\mathrm{TF}}, \tilde V_\ell^{\mathrm{AR}}\right]
=
Q_\ell^{\mathrm{Aug}} R_\ell^{\mathrm{Aug}}.
\end{equation}
Here, $Q_\ell^{\mathrm{Aug}}$ contains the orthonormal basis vectors spanning the Augmented HalluSpace.
For notational consistency with the HalluSpace bases, we denote this orthonormal basis as
\begin{equation}
V_\ell^{\mathrm{Aug}} := Q_\ell^{\mathrm{Aug}} .
\end{equation}
Finally, DARE edits the FFN down-projection weights of the LVLM block using the corresponding null-space projection as follows:
\begin{equation}
P_\ell = I - V_\ell^{\mathrm{Aug}} {V_\ell^{\mathrm{Aug}}}^{\top},
\end{equation}
\begin{equation}
W_{\ell,\mathrm{down}}^{\mathrm{edit}} =
P_\ell W_{\ell,\mathrm{down}}^{\mathrm{orig}}.
\end{equation}

\noindent By applying this orthonormalized projection to the FFN down-projection layer, DARE suppresses hallucination-related representation directions while preserving the overall transformer architecture and inference pipeline.

\subsection{Visual-Difference Editing}
\label{sec:visual_editing}

\noindent The analysis in Sec.~\ref{sec:introduction} reveals that hallucination editing based solely on textual contrasts can unintentionally perturb visual components of the multimodal representation. 
As illustrated in Fig.~\ref{fig:motivation}(C), editing directions derived from hallucinated and truthful textual responses are accompanied by reduced attention to visual tokens across layers.
This observation suggests that textual contrasts alone may not sufficiently capture the visual factors that contribute to hallucinated generations. To explicitly incorporate visual evidence into the editing process, we introduce \textit{Visual-difference editing}, which constructs hallucination directions from controlled visual contrasts. We utilize the BEAF dataset~\cite{ye2025beaf}, which provides paired images derived from MSCOCO~\cite{lin2014microsoft} where specific objects are naturally removed from the original image.
Each pair therefore forms a controlled visual counterfactual: one image contains the object while the other does not.
Using MSCOCO annotations, we associate each pair with object-level ground-truth descriptions that reflect whether the removed object should appear in the caption. To isolate representation differences that arise purely from visual changes, hidden representations are extracted under teacher-forcing.
Teacher-forcing ensures that both images are processed under the same textual sequence, allowing representation differences to reflect variations in visual input rather than differences in generated language. This design is crucial because autoregressive decoding may produce different token sequences for the two images, which would entangle visual effects with generation dynamics. By aligning the textual context through teacher-forcing, we obtain a controlled representation contrast that isolates visual differences. Formally, we construct paired samples $\mathcal{D}_{vis}=\{(x_i^{H},x_i^{T})\}_{i=1}^{N}$, 
where $x_i^{H}$ denotes the hallucinated condition and $x_i^{T}$ denotes the truthful condition aligned with the visual content. Running teacher-forcing through the LVLM yields hidden representations at layer $\ell$. 
After pooling over token positions, the resulting representations are stacked into matrices 
$X_{\ell}^{H}, X_{\ell}^{T} \in \mathbb{R}^{N \times D}$. We then define the visual contrast matrix as 
\begin{equation}
    E_{\ell}^{vis}=X_{\ell}^{H}-X_{\ell}^{T},
\end{equation}
which captures representation differences induced by controlled visual changes. We extract the dominant directions by decomposing $E_{\ell}^{vis}$ via SVD as
\begin{equation}
E_{\ell}^{vis}=U_{\ell}\Sigma_{\ell}(V_{\ell}^{vis})^{\top}.
\end{equation}

\noindent The top-$k_{vis}$ right singular vectors are selected to construct the visual-difference HalluSpace, denoted as 
$V_{\ell,k_{vis}}^{vis}=[v_1^{vis},\dots,v_{k_{vis}}^{vis}]$. 
Finally, the attention output projection weights are edited through null-space projection as
\begin{equation}
W_{\ell,o}^{edit}
=
\left(I-V_{\ell,k_{vis}}^{vis}(V_{\ell,k_{vis}}^{vis})^{\top}\right)W_{\ell,o}^{orig}.
\end{equation}

\noindent Editing the attention output projection allows DARE to directly influence how visual features produced by the attention module are injected into the residual stream. This enables the editing process to suppress hallucination-related representations while preserving visually grounded information within the multimodal representation space.
\section{Experiments}
\label{sec:experiments}

\noindent In this section, we evaluated our method against state-of-the-art approaches for mitigating object hallucination in LVLMs. We first describe the experimental setup in Sec.~\ref{sec4.1}, followed by quantitative comparisons with existing methods in Sec.~\ref{sec4.2}. Additional analyses and ablation studies are presented in Sec.~\ref{sec4.3}.

\subsection{Experimental Setup}
\label{sec4.1}
\noindent\textbf{Models.} We conducted experiments on three widely used open-source LVLMs: 
LLaVA-1.5~\cite{liu2023improved}, MiniGPT-4~\cite{zhu2023minigpt4}, and mPLUG-Owl2~\cite{ye2024mplug}. All experiments are performed using the official released checkpoints without additional fine-tuning. Unless otherwise specified, our primary analysis focuses on LLaVA-1.5.

\noindent\textbf{Baselines.}
We compared our method with several representative approaches for mitigating hallucinations in LVLMs. \textit{Standard decoding baselines.}
Greedy decoding and Beam Search~\cite{freitag2017beam} are included as standard generation baselines, reflecting model behavior without explicit hallucination mitigation. \textit{Decoding-based methods.}
VCD~\cite{leng2024mitigating}, OPERA~\cite{huang2024opera}, ONLY~\cite{wan2025only}, and CMI-VLD~\cite{fang2025grounding} mitigate hallucination by modifying token selection during decoding or introducing inference-time constraints. \textit{Representation intervention methods.}
SID~\cite{huo2025selfintrospective} perturbs intermediate vision representations and subtracts amplified hallucinated logits to encourage more factual predictions. VAF~\cite{yin2025clearsight} reweights cross-attention scores in multimodal fusion layers to strengthen visual signals while suppressing system-prompt attention. \textit{Model editing method.} Nullu~\cite{yang2025nullu} mitigates hallucinations by projecting model weights onto the null space of hallucination-related components, thereby removing hallucination-inducing directions without additional training.
\par\smallskip
\noindent\textbf{Benchmarks.}
We evaluated hallucination mitigation performance using two widely adopted benchmarks for object hallucination: CHAIR~\cite{rohrbach2018object} and POPE~\cite{li2023evaluating}. 
Both benchmarks use images from the MSCOCO validation set~\cite{lin2014microsoft}. 
CHAIR measures the proportion of non-existent objects mentioned in generated captions, whereas POPE evaluates factual consistency through yes/no questions about object presence in images. 
To further assess general multimodal capabilities, we additionally reported results on MME~\cite{fu2023mme}, which evaluates perception and reasoning abilities across diverse multimodal tasks.

\noindent\textbf{Implementation Details.}
All baseline implementations follow the official configurations provided in the HuggingFace Transformers repository. 
Additional implementation details are provided in the Supplementary Material.

\subsection{Experimental Results}\label{sec4.2}
\begin{table*}[!t]
\centering
\setlength{\tabcolsep}{1pt}
\caption{The \textbf{CHAIR} evaluation results of three baseline LVLMs with existing OH mitigation methods. The down-arrow ($\downarrow$) indicates that lower $\text{\textbf{CHAIR}}_S$ and $\text{\textbf{CHAIR}}_I$ result in fewer object hallucinations, while the up-arrow ($\uparrow$) indicates that higher \textbf{BLEU} result in better general generation quality. Bold and underlined indicates the best and the second best, respectively. We used 64 as the max token length.}
\resizebox{\linewidth}{!}{
\begin{tabular}{ l | c c c | c c c | c c c}
\specialrule{1.2pt}{0pt}{1.2pt}
\multirow{2}{*}{\textbf{Method}} 
&\multicolumn{3}{c|}{\textbf{LLaVA-1.5}}
&\multicolumn{3}{c|}{\textbf{MiniGPT-4}}
&\multicolumn{3}{c}{\textbf{mPLUG-Owl2}} \\
\cmidrule{2-10}
&\textbf{CHAIR}$_S \downarrow$ &\textbf{CHAIR}$_I \downarrow$ & \textbf{BLEU} $\uparrow$ 
&\textbf{CHAIR}$_S \downarrow$ &\textbf{CHAIR}$_I \downarrow$ & \textbf{BLEU} $\uparrow$ 
&\textbf{CHAIR}$_S \downarrow$ &\textbf{CHAIR}$_I \downarrow$ & \textbf{BLEU} $\uparrow$   \\ 
\midrule
Greedy
&$\text{24.32}{\pm \text{2.92}}$ &$\text{8.91}{\pm \text{0.46}}$ &$\text{15.16}{\pm \text{0.30}}$ 
&$\text{35.91}{\pm \text{3.15}}$ &$\text{14.61}{\pm \text{0.09}}$ &$\text{14.43}{\pm \text{0.46}}$ 
&$\text{24.96}{\pm \text{1.43}}$ &$\text{9.16}{\pm \text{1.04}}$ &$\text{14.74}{\pm \text{0.62}}$   
\\
Beam \cite{freitag2017beam}
&$\text{22.61}{\pm \text{1.64}}$ &$\text{7.01}{\pm \text{0.95}}$ &$\text{16.17}{\pm \text{0.13}}$ 
&$\text{27.33}{\pm \text{2.08}}$ &$\text{11.11}{\pm \text{0.98}}$ &$\text{15.09}{\pm \text{0.21}}$ 
&$\text{21.80}{\pm \text{1.31}}$ &$\text{7.09}{\pm \text{0.54}}$ &$\text{16.17}{\pm \text{0.10}}$  
\\
VCD \cite{leng2024mitigating}
&$\text{21.15}{\pm \text{1.25}}$ &$\text{7.19}{\pm \text{0.26}}$ &$\text{14.91}{\pm \text{0.46}}$ 
&$\text{30.60}{\pm \text{1.95}}$ &$\text{11.06}{\pm \text{1.11}}$ &$\text{14.94}{\pm \text{0.22}}$ 
&$\text{20.56}{\pm \text{2.40}}$ &$\text{7.01}{\pm \text{0.34}}$ &$\text{15.46}{\pm \text{0.19}}$  
\\
OPERA \cite{huang2024opera}
&$\text{17.61}{\pm \text{1.16}}$ &$\underline{\text{6.09}{\pm \text{1.11}}}$ &$\text{15.21}{\pm \text{0.49}}$ 
&$\text{29.64}{\pm \text{0.18}}$ &$\text{10.91}{\pm \text{0.76}}$ &$\text{15.42}{\pm \text{0.52}}$ 
&$\text{20.19}{\pm \text{0.15}}$ &$\text{7.16}{\pm \text{0.13}}$ &$\text{15.31}{\pm \text{0.64}}$  
\\
SID \cite{huo2025selfintrospective}
&$\text{17.95}{\pm \text{1.93}}$ &$\text{6.64}{\pm \text{0.11}}$ &$\text{15.04}{\pm \text{0.24}}$ 
&$\text{27.16}{\pm \text{2.06}}$ &$\text{10.34}{\pm \text{0.46}}$ &$\text{15.83}{\pm \text{0.27}}$ 
&$\text{19.66}{\pm \text{1.43}}$ &$\text{7.13}{\pm \text{0.18}}$ &$\text{15.94}{\pm \text{0.16}}$  
\\
VAF \cite{yin2025clearsight}
&$\text{17.92}{\pm \text{1.06}}$ &$\text{7.14}{\pm \text{1.05}}$ &$\text{15.01}{\pm \text{0.62}}$ 
&$\text{25.61}{\pm \text{2.96}}$ &$\text{10.64}{\pm \text{0.43}}$ &$\text{15.34}{\pm \text{0.26}}$ 
&$\text{18.66}{\pm \text{0.15}}$ &$\text{7.09}{\pm \text{0.64}}$ &$\text{14.94}{\pm \text{0.39}}$  
\\
Nullu \cite{yang2025nullu}
&${\text{20.53}{\pm \text{1.40}}}$ &${\text{7.03}{\pm \text{0.86}}}$ &${\text{15.37}{\pm \text{0.18}}}$ 
&${\text{23.93}{\pm \text{0.31}}}$ &${\text{10.40}{\pm \text{0.12}}}$ &${\text{14.68}{\pm \text{0.24}}}$ 
&${\text{18.80}{\pm \text{0.72}}}$ &$\underline{{\text{6.86}{\pm \text{0.30}}}}$ &${\text{16.02}{\pm \text{0.24}}}$ 
\\
ONLY \cite{wan2025only}
&$\text{21.64}{\pm \text{0.64}}$ &$\text{6.71}{\pm \text{0.67}}$ &$\text{15.29}{\pm \text{0.36}}$ 
&$\text{24.17}{\pm \text{2.48}}$ &$\text{9.07}{\pm \text{0.93}}$ &$\text{14.94}{\pm \text{0.67}}$ 
&$\text{19.43}{\pm \text{1.18}}$ &$\text{7.37}{\pm \text{1.08}}$ &$\text{14.42}{\pm \text{0.71}}$  
\\
CMI-VLD \cite{fang2025grounding}
&$\underline{\text{16.42}{\pm \text{0.75}}}$ &$\text{6.59}{\pm \text{0.81}}$ &$\text{15.40}{\pm \text{0.86}}$ 
&$\underline{\text{23.80}{\pm \text{1.73}}}$ &$\underline{\text{8.90}{\pm \text{0.53}}}$ &$\text{15.38}{\pm \text{0.33}}$ 
&$\underline{\text{17.76}{\pm \text{0.84}}}$ &$\text{6.95}{\pm \text{0.94}}$ &$\text{15.11}{\pm \text{0.34}}$ 
\\
\midrule
\rowcolor{lightgray!40} \textbf{DARE (Ours)} 
&${\textbf{14.93} {\pm \textbf{1.68}}}$ & ${\textbf{6.07} {\pm \textbf{0.82}}}$ & $\text{15.61}{\pm \text{0.59}}$
&${\textbf{22.64} {\pm \textbf{1.49}}}$ & ${\textbf{8.73} {\pm \textbf{0.96}}}$ & $\text{15.53}{\pm \text{0.34}}$
&${\textbf{16.81} {\pm \textbf{0.43}}}$ & ${\textbf{6.57} {\pm \textbf{0.84}}}$ & $\text{16.02}{\pm \text{0.19}}$
\\
\specialrule{1.2pt}{0pt}{1.2pt}
\end{tabular}
}
\label{tab:chair}
\end{table*}

\noindent\textbf{CHAIR Results.}
Table~\ref{tab:chair} shows that DARE consistently achieves the lowest hallucination scores across all evaluated LVLMs.
On LLaVA-1.5, DARE reduces CHAIR$_S$ from 24.32 to 14.93 and CHAIR$_I$ from 8.91 to 6.07, showing substantial improvements over both the base model and prior mitigation methods.
Similar trends are observed on MiniGPT-4 and mPLUG-Owl2, indicating that the proposed editing strategy is not limited to a single backbone.
Unlike TF-based editing methods such as Nullu~\cite{yang2025nullu}, which derive hallucination directions from static teacher-forced contrasts, DARE incorporates AR-aware transitions that directly reflect generation-time dynamics.
This allows DARE to better suppress hallucinated object mentions during actual autoregressive decoding while maintaining comparable generation quality, as reflected by the BLEU scores.

\begin{table*}[!ht]
\centering
\setlength{\tabcolsep}{1.0pt}
\caption{The \textbf{POPE} with random sampling evaluation results on MSCOCO of three baseline LVLMs with existing OH mitigation methods. The up-arrow ($\uparrow$) indicates that higher \textbf{Accuracy}, \textbf{Precision} and \textbf{F1 Score} result in better performance. Bold and underlined indicates the best and the second best, respectively. We used 64 as the max token number in this experiment. Results for random, popular and adversarial sampling, including recall metric are in the supplementary.}
\resizebox{\linewidth}{!}{
\begin{tabular}{ l | c c c | c c c | c c c}
\specialrule{1.2pt}{0pt}{1.2pt}
\multirow{2}{*}{\textbf{Method}} 
&\multicolumn{3}{c|}{\textbf{LLaVA-1.5}}
&\multicolumn{3}{c|}{\textbf{MiniGPT-4}}
&\multicolumn{3}{c}{\textbf{mPLUG-Owl2}} \\
\cmidrule{2-10}
&\textbf{Accuracy} $\uparrow$ &\textbf{Precision} $\uparrow$ &\textbf{F1 Score} $\uparrow$ 
&\textbf{Accuracy} $\uparrow$ &\textbf{Precision} $\uparrow$ &\textbf{F1 Score} $\uparrow$ 
&\textbf{Accuracy} $\uparrow$ &\textbf{Precision} $\uparrow$ &\textbf{F1 Score} $\uparrow$   \\ 
\midrule
Greedy
&$\text{79.53}{\pm \text{1.56}}$ &$\text{82.64}{\pm \text{0.84}}$ &$\text{82.47}{\pm \text{0.71}}$ 
&$\text{72.84}{\pm \text{1.41}}$ &$\text{70.79}{\pm \text{1.15}}$ &$\text{69.68}{\pm \text{0.93}}$ 
&$\text{75.71}{\pm \text{0.90}}$ &$\text{77.68}{\pm \text{1.35}}$ &$\text{80.71}{\pm \text{0.48}}$   
\\
Beam \cite{freitag2017beam}
&$\text{79.57}{\pm \text{0.51}}$ &$\text{83.04}{\pm \text{0.81}}$ &$\text{80.49}{\pm \text{0.85}}$ 
&$\text{72.70}{\pm \text{1.40}}$ &$\text{71.27}{\pm \text{0.49}}$ &$\text{70.76}{\pm \text{0.34}}$ 
&$\text{75.97}{\pm \text{1.09}}$ &$\text{75.94}{\pm \text{1.04}}$ &$\text{80.78}{\pm \text{0.27}}$  
\\
VCD \cite{leng2024mitigating}
&$\text{81.61}{\pm \text{0.29}}$ &$\text{82.90}{\pm \text{0.95}}$ &$\text{82.79}{\pm \text{1.11}}$ 
&$\text{75.62}{\pm \text{0.46}}$ &$\text{75.06}{\pm \text{0.77}}$ &$\text{78.17}{\pm \text{0.85}}$ 
&$\text{78.48}{\pm \text{1.30}}$ &$\text{79.34}{\pm \text{0.91}}$ &$\text{83.62}{\pm \text{0.55}}$  
\\
OPERA \cite{huang2024opera}
&$\text{87.83}{\pm \text{1.31}}$ &$\text{88.34}{\pm \text{0.84}}$ &$\text{88.70}{\pm \text{0.90}}$ 
&$\text{75.46}{\pm \text{0.33}}$ &$\text{75.60}{\pm \text{0.18}}$ &$\text{79.30}{\pm \text{0.39}}$ 
&$\text{78.03}{\pm \text{0.59}}$ &$\text{81.90}{\pm \text{0.47}}$ &$\text{84.03}{\pm \text{1.64}}$  
\\
SID \cite{huo2025selfintrospective}
&$\text{87.90}{\pm \text{1.32}}$ &$\text{86.64}{\pm \text{0.47}}$ &$\text{88.03}{\pm \text{0.33}}$ 
&$\text{76.41}{\pm \text{0.67}}$ &$\text{79.67}{\pm \text{0.08}}$ &$\text{79.45}{\pm \text{0.48}}$ 
&$\text{78.26}{\pm \text{0.80}}$ &$\text{82.39}{\pm \text{1.10}}$ &$\text{84.63}{\pm \text{0.79}}$  
\\
VAF \cite{yin2025clearsight}
&$\text{87.96}{\pm \text{0.47}}$ &$\text{88.16}{\pm \text{1.24}}$ &$\text{89.07}{\pm \text{0.22}}$ 
&$\text{78.90}{\pm \text{2.08}}$ &$\text{78.11}{\pm \text{0.30}}$ &$\text{79.46}{\pm \text{0.80}}$ 
&$\text{78.05}{\pm \text{1.38}}$ &$\text{84.46}{\pm \text{0.81}}$ &$\text{83.29}{\pm \text{0.54}}$  
\\
Nullu \cite{yang2025nullu}
&$\underline{{\text{88.73}{\pm \text{1.43}}}}$ &${\text{88.70}{\pm \text{0.92}}}$ &${\text{87.56}{\pm \text{0.64}}}$ 
&${\text{79.67}{\pm \text{0.94}}}$ &${\text{79.50}{\pm \text{0.37}}}$ &${\text{80.09}{\pm \text{0.47}}}$ 
&${\text{79.38}{\pm \text{0.75}}}$ &${\text{83.67}{\pm \text{0.61}}}$ &${\text{85.41}{\pm \text{1.03}}}$ 
\\
ONLY \cite{wan2025only}
&$\text{88.64}{\pm \text{0.91}}$ &$\text{88.41}{\pm \text{0.74}}$ &$\text{88.70}{\pm \text{0.70}}$ 
&$\text{79.37}{\pm \text{1.43}}$ &$\text{80.09}{\pm \text{0.59}}$ &$\text{80.26}{\pm \text{0.12}}$ 
&$\underline{\text{81.66}{\pm \text{0.90}}}$ &$\text{84.17}{\pm \text{1.10}}$ &$\text{84.64}{\pm \text{1.37}}$  
\\
CMI-VLD \cite{fang2025grounding}
&$\text{88.01}{\pm \text{0.17}}$ &$\underline{\text{89.64}{\pm \text{0.67}}}$ &$\underline{\text{89.17}{\pm \text{1.26}}}$ 
&$\underline{\text{80.18}{\pm \text{0.81}}}$ &$\underline{\text{81.74}{\pm \text{0.55}}}$ &$\underline{\text{80.61}{\pm \text{0.77}}}$ 
&$\text{81.59}{\pm \text{0.21}}$ &$\underline{\text{85.19}{\pm \text{0.81}}}$ &$\underline{\text{85.70}{\pm \text{1.28}}}$ 
\\
\midrule
\rowcolor{lightgray!40} \textbf{DARE (Ours)}
&${\textbf{89.74} {\pm \textbf{0.54}}}$ & ${\textbf{91.40} {\pm \textbf{0.71}}}$ & ${\textbf{90.04} {\pm \textbf{0.37}}}$
&${\textbf{80.96} {\pm \textbf{0.49}}}$ & ${\textbf{81.97} {\pm \textbf{0.40}}}$ & ${\textbf{81.67} {\pm \textbf{0.75}}}$
&${\textbf{82.94} {\pm \textbf{1.07}}}$ & ${\textbf{85.36} {\pm \textbf{0.18}}}$ & ${\textbf{85.78} {\pm \textbf{0.48}}}$
\\
\specialrule{1.2pt}{0pt}{1.2pt}
\end{tabular}
}
\label{tab:pope}
\end{table*}

\noindent\textbf{POPE Results.}
Table~\ref{tab:pope} further evaluates object-existence reasoning through yes/no questions.
DARE consistently improves accuracy, precision, and F1 score across the evaluated LVLMs, demonstrating that its gains are not restricted to caption generation.
Since POPE primarily measures whether models incorrectly affirm the presence of non-existent objects, these improvements indicate that DARE effectively reduces false-positive object predictions.
Compared with TF-based editing, which captures representation differences under teacher forcing, DARE accounts for the model's actual autoregressive generation process. This leads to more reliable alignment between textual responses and visual evidence.

\noindent\textbf{MME Results.}
Fig.~\ref{fig:mme} evaluates whether hallucination mitigation comes at the cost of general multimodal capability.
DARE maintains competitive or improved MME performance compared with the base model and strong hallucination mitigation baselines.
Since MME covers diverse perception and reasoning skills, including object existence, counting, position, color, OCR, and common-sense reasoning, these results suggest that DARE does not simply suppress object mentions or weaken visual understanding.
Instead, DARE reduces hallucination while preserving the model's broader multimodal perception ability.
This supports the design of combining TF/AR-based hallucination suppression with visual-difference editing, which aims to mitigate hallucinated generations without disrupting visually grounded representations.

\begin{table}[!t]
	\begin{minipage}[!t]{0.49\linewidth}
        \includegraphics[width=\textwidth]{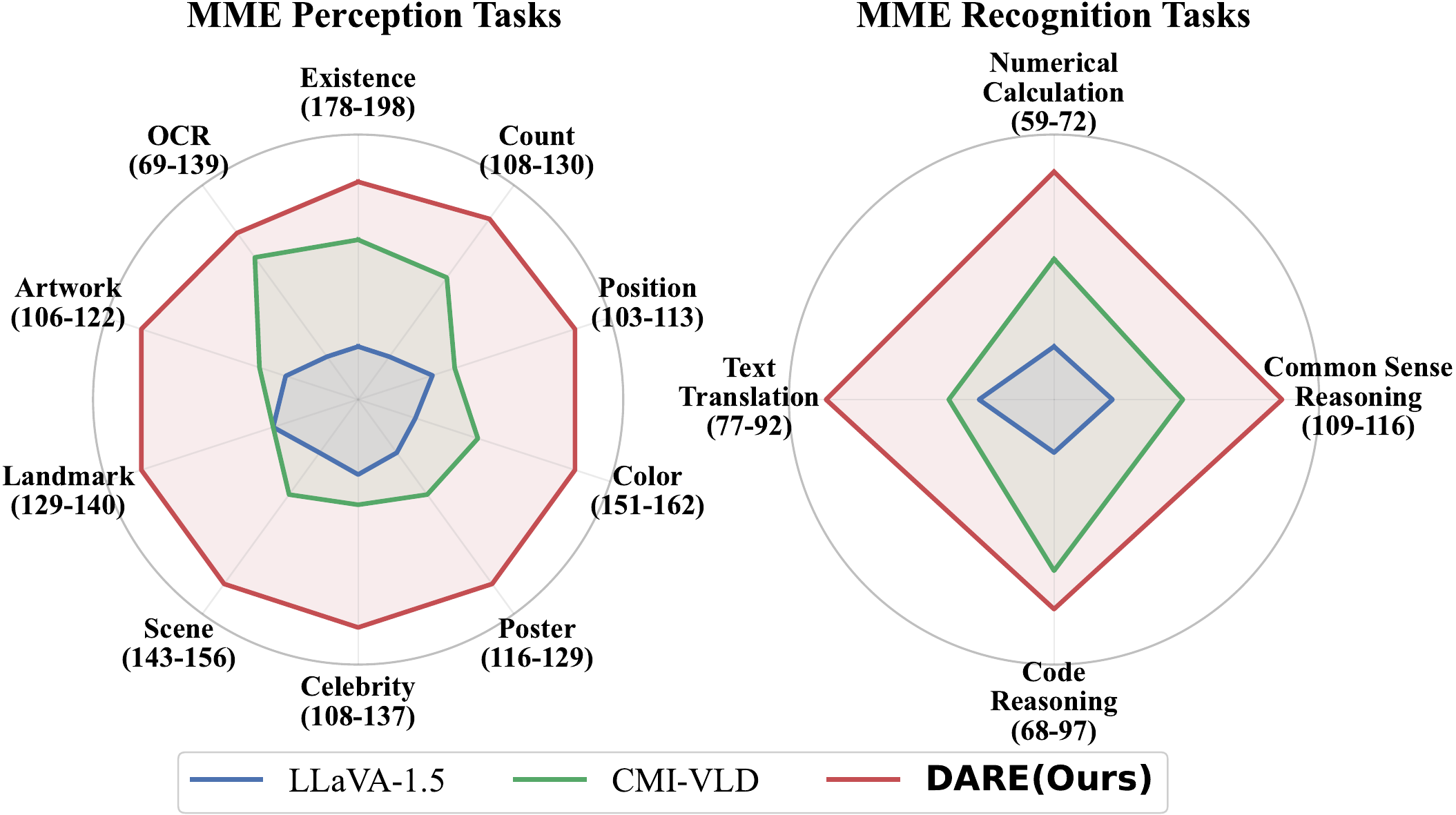}
        \captionof{figure}{Radar visualization of MME full-set evaluation results.}
        \label{fig:mme}
	\end{minipage}
	\begin{minipage}[!t]{0.49\linewidth}
        \caption{Ablation study of the proposed DARE components.}
        \renewcommand{\arraystretch}{0.8}
        \renewcommand{\tabcolsep}{0.8mm}
        \resizebox{\textwidth}{!}{
            \begin{tabular}{lccc|cc}
            \specialrule{1.2pt}{0pt}{1.2pt}
            \multirow{2}{*}{Method} & TF & AR & Visual & \multirow{2}{*}{\textbf{CHAIR}$_S \downarrow$} & \multirow{2}{*}{\textbf{CHAIR}$_I \downarrow$} \\
            & Editing & Editing & Diff. &&  \\
            \midrule
            Base                &  &   &  & $\text{24.32}{\pm \text{2.92}}$ &$\text{8.91}{\pm \text{0.46}}$  \\
            \midrule
            (A)          & \checkmark &  &  & ${\text{20.53}{\pm \text{1.40}}}$ &${\text{7.03}{\pm \text{0.86}}}$  \\
            (B)          &  & \checkmark &  &  ${\text{17.61}{\pm \text{1.64}}}$ &${\text{6.94}{\pm \text{1.06}}}$  \\
            (C)      &  &  & \checkmark &  ${\text{17.81}{\pm \text{1.28}}}$ &${\text{7.20}{\pm \text{0.73}}}$  \\
            (D)      & \checkmark & \checkmark &  & ${\text{15.83}{\pm \text{1.48}}}$ & ${\text{6.47}{\pm \text{1.03}}}$  \\
            (E)      &  & \checkmark & \checkmark &  ${\text{15.57}{\pm \text{1.70}}}$ &${\text{6.41}{\pm \text{0.83}}}$  \\
            (F)      & \checkmark &  & \checkmark &  ${\text{17.17}{\pm \text{1.22}}}$ &${\text{6.62}{\pm \text{0.50}}}$  \\
            \midrule
            \textbf{DARE (Ours)}       & \checkmark & \checkmark & \checkmark & ${\textbf{14.93} {\pm \textbf{1.68}}}$ & ${\textbf{6.07} {\pm \textbf{0.82}}}$   \\
            \specialrule{1.2pt}{0pt}{1.2pt}
            \end{tabular}
        }
		\label{tab:ablation}
	\end{minipage}
\end{table}

\subsection{Ablation Studies and Further Analysis}\label{sec4.3}

\noindent\textbf{Component Ablation.}
Table~\ref{tab:ablation} presents an ablation study of the proposed DARE components.
Each individual component improves over the base model: TF-based editing (A), AR-aware editing (B), and visual-difference editing (C) reduce both CHAIR$_S$ and CHAIR$_I$, indicating that each provides a useful hallucination-mitigation signal.
Among them, AR-aware editing achieves the strongest single-component performance, suggesting the importance of modeling hallucination directions that emerge during auto-regressive generation. Pairwise combinations further improve performance.
TF+AR editing (D) outperforms either TF or AR alone, showing that teacher-forced and generation-aware hallucination signals are complementary.
AR+visual-difference editing (E) gives the best pairwise result, while TF+visual-difference editing (F) also improves over TF alone, confirming that visual-difference cues help preserve image-grounded information during editing.
Finally, combining all three components achieves the best overall performance, reducing CHAIR$_S$ from 24.32 to 14.93 and CHAIR$_I$ from 8.91 to 6.07.
These results demonstrate that TF-based signals, AR generation dynamics, and visual-difference information provide complementary cues for effective hallucination mitigation.

\begin{table}[!t]
\centering
\setlength{\tabcolsep}{1pt}
\caption{Effects of editing layers \{ $\ell$ \}, rank $k$ and window size for constructing AR-HalluSpace.}
\resizebox{\linewidth}{!}{
\begin{tabular}{ c | c c | c | c c | c | c c }
\specialrule{1.2pt}{0pt}{1.2pt}
\textbf{\{ $\ell$ \}} & \textbf{CHAIR}$_S \downarrow$ & \textbf{CHAIR}$_I \downarrow$  & \textbf{$\,\,\, k \,\,\,$} & \textbf{CHAIR}$_S \downarrow$ & \textbf{CHAIR}$_I \downarrow$ & \textbf{Window} & \textbf{CHAIR}$_S \downarrow$ & \textbf{CHAIR}$_I \downarrow$ \\
\midrule 
\textbf{16-31} & \textbf{15.11} & \textbf{6.59} & 1 & 18.44 & 7.80 & 1 & 16.92 & 6.87  \\
24-31 & 16.87 & 6.93 & \textbf{2} & \textbf{15.11} & \textbf{6.59} & 2 & 16.15 & 6.82  \\
0-15 & 17.70 & 7.22 & 4 & 17.36 & 6.90 & \textbf{3} & \textbf{15.11} & \textbf{6.59}  \\
0-7 & 17.37 & 7.18 & 8 & 17.09 & 6.83 & 4 & 15.84 & 6.63  \\
\specialrule{1.2pt}{0pt}{1.2pt}
\end{tabular}
}
\label{tab:hyperparams}
\end{table}

\par\smallskip
\noindent\textbf{Hyperparameter Sensitivity.}
Table~\ref{tab:hyperparams} analyzes the sensitivity of AR-aware editing to the edited layers, HalluSpace rank, and temporal window size.
Editing higher layers (16--31) performs best, suggesting that hallucination-related directions are more effectively suppressed near the language generation stage.
For the rank, $k=2$ achieves the lowest CHAIR scores, while larger ranks slightly degrade performance, indicating that AR hallucination signals lie in a compact low-dimensional subspace.
For the temporal window, a window size of 3 gives the best result, suggesting that hallucination-inducing transitions are concentrated within a short context before the hallucinated token.

\begin{table}[!t]
\centering
\caption{Generality evaluation on recent LVLMs with different model scales. DARE is evaluated on Qwen3-VL and Gemma-3 across 4B, 8B, and 12B variants, showing consistent improvements over the base model and Nullu on both CHAIR and POPE.
}
\label{tab:recent_lvlm}
\setlength{\tabcolsep}{1.0pt}
\renewcommand{\arraystretch}{1.2}
\begin{tabular}{@{}l|c|c|c|c@{}}
\specialrule{1.2pt}{0pt}{1.2pt}
\multicolumn{5}{c}{\textbf{CHAIR Evaluation} $(C_S / C_I)\downarrow$} \\
\hline
Method & Qwen3-VL-4B & Qwen3-VL-8B & Gemma-3-4B & Gemma-3-12B \\
\hline
Base  & 39.20 / 8.83 & 36.40 / 8.38 & 32.60 / 7.86 & 33.40 / 6.97 \\
Nullu & 37.80 / 8.44 & 34.80 / 7.48 & 30.50 / 7.01 & 32.56 / 6.29 \\
\hline
\textbf{DARE (Ours)}
& \textbf{36.75 / 7.80}
& \textbf{33.17 / 6.51}
& \textbf{29.67 / 6.78}
& \textbf{30.91 / 6.08} \\
\hline
\multicolumn{5}{c}{\textbf{POPE Evaluation} $(\mathrm{Acc.} / \mathrm{F1})\uparrow$} \\
\hline
Method & Qwen3-VL-4B & Qwen3-VL-8B & Gemma-3-4B & Gemma-3-12B \\
\hline
Base  & 89.99 / 89.40 & 88.88 / 88.22 & 84.30 / 84.53 & 84.33 / 85.28 \\
Nullu & 90.31 / 89.91 & 88.93 / 88.07 & 84.88 / 84.56 & 85.16 / 85.49 \\
\hline
\textbf{DARE (Ours)}
& \textbf{90.84 / 90.66}
& \textbf{89.67 / 89.20}
& \textbf{86.10 / 87.46}
& \textbf{86.73 / 87.08} \\
\specialrule{1.2pt}{0pt}{1.2pt}
\end{tabular}
\end{table}

\par\smallskip
\noindent\textbf{Generality on Recent LVLMs.}
To further examine whether DARE generalizes beyond the LVLMs used in our main comparison, we additionally evaluate recent model families and scales, including Qwen3-VL-4B~\cite{bai2025qwen3}, Qwen3-VL-8B, Gemma-3-4B~\cite{gemmateam2025gemma3technicalreport}, and Gemma-3-12B.
As shown in Table~\ref{tab:recent_lvlm}, DARE consistently improves over both the base model and Nullu on CHAIR and POPE across all evaluated backbones and model sizes.
These results indicate that DARE is not limited to a specific LVLM family or parameter scale, and provides consistent gains for both caption-level hallucination and object-existence reasoning in recent LVLMs.
\par\smallskip
\noindent\textbf{Throughput--Hallucination Trade-off.}
Fig.~\ref{fig:throughput} compares generation throughput and CHAIR$_S$ across hallucination mitigation methods.
Decoding-based methods such as VCD and OPERA reduce hallucinations but require additional decoding operations, leading to lower throughput.
In contrast, DARE achieves stronger hallucination mitigation while maintaining nearly the same generation speed as efficient editing-based methods.
Compared with Nullu, DARE further reduces hallucination without introducing inference-time overhead, since the model is edited once through null-space projection and then decoded with the original inference pipeline.
Thus, DARE provides a favorable balance between hallucination reduction and generation efficiency.

\begin{table}[!t]
\begin{minipage}[!t]{.45\textwidth}
\centering
     \includegraphics[width=\textwidth]{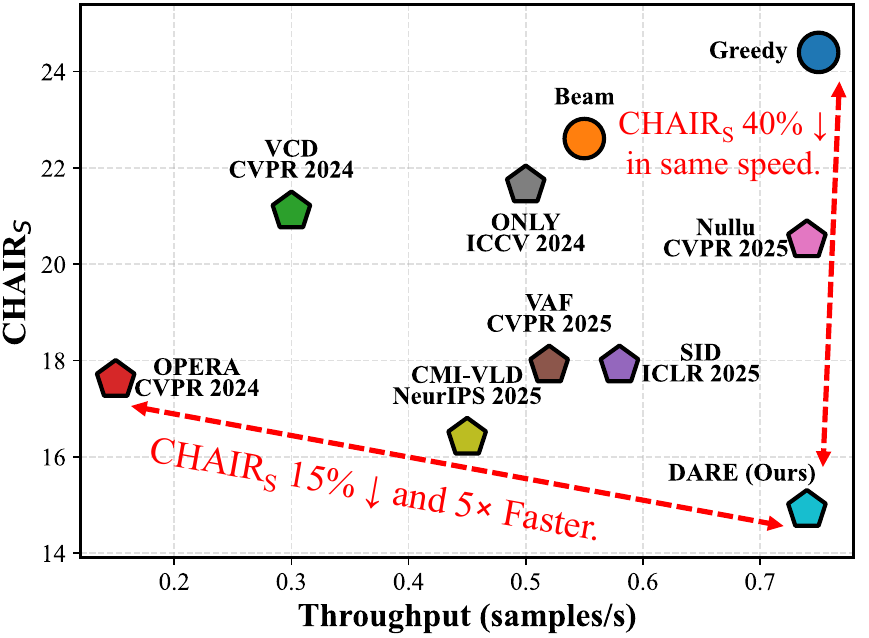}
     \vspace{1mm}
     \captionof{figure}{Throughput–Hallucination Trade-off. Inference throughput and hallucination performance are measured on an NVIDIA RTX 4090.}
\label{fig:throughput}
\end{minipage}
\hspace{2mm}
\begin{minipage}[!t]{.55\textwidth}
\centering
     \includegraphics[width=\textwidth]{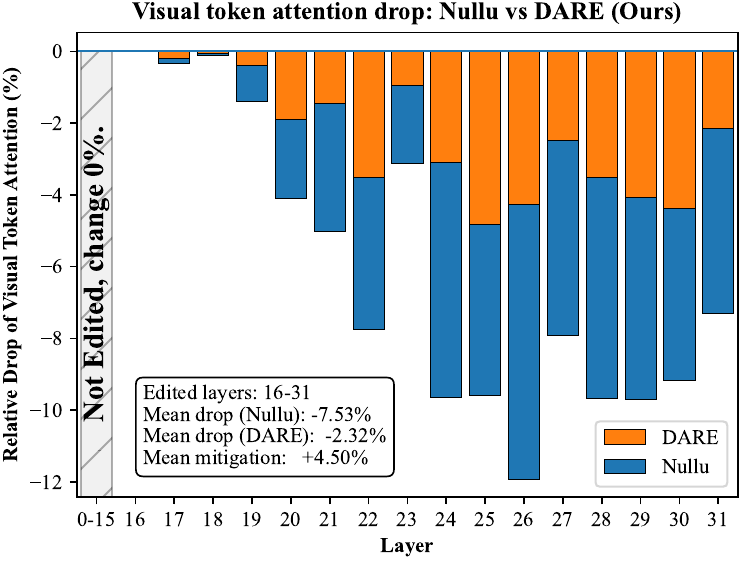}
     \captionof{figure}{Visual attention drop after proposed augmented HalluSpace and Visual-difference HalluSpace editing.}
\label{fig:visdrop}
\end{minipage}
\end{table}

\par\smallskip
\noindent\textbf{Visual Attention Drop after Editing.}
Fig.~\ref{fig:visdrop} analyzes how editing affects visual-token attention.
Consistent with Fig.~\ref{fig:motivation}(C), TF-based editing causes a noticeable drop in visual attention, suggesting that text-derived hallucination directions can also perturb multimodal interactions. In contrast, DARE mitigates this drop: the average visual-token attention drop over layers 16--31 decreases from $-7.53\%$ for Nullu to $-2.32\%$ for DARE. The reduced attention drop indicates that DARE better preserves visual-token interactions during editing. This behavior is consistent with our design: TF and AR directions are integrated into the Augmented HalluSpace to suppress hallucination-related generation signals, while visual-difference editing complements them under visual-token disruption. 
\section{Conclusion}
\label{sec:conclusion}

\noindent We presented DARE (Dual-path Auto-Regressive-aware Editing), a training-free weight editing framework for mitigating object hallucination in LVLMs. Motivated by our analyses of TF-based editing, we showed that teacher-forced representations can retain strong evidence for truthful tokens, yet TF-based editing often yields limited changes in the truthful--hallucinated decision margin under autoregressive decoding and can be accompanied by reduced attention to visual tokens. To address these gaps, DARE integrates dual-path contrasts---textual contrasts and image contrasts---with AR-aware signals derived from decoding trajectories. Concretely, DARE edits the FFN down-projection using an Augmented HalluSpace that combines TF and AR directions, and applies visual-difference editing on the attention output projection to better preserve visually grounded information flow. Extensive experiments on standard LVLM hallucination benchmarks demonstrated that DARE consistently reduces hallucinations while maintaining competitive multimodal performance and inference efficiency, since the editing is performed once without decoding-time overhead. Additional evaluations on recent LVLM families with different model scales further show that DARE is not limited to a specific backbone or parameter size. We hope DARE serves as a practical foundation for hallucination mitigation that goes beyond teacher-forced analysis and reflects autoregressive generation behavior. 
\clearpage
\section*{Acknowledgements}
This work was supported by Institute of Information and Communications Technology Planning and Evaluation(IITP) grant funded by the Korea government(MSIT) (RS-2019-II190004, Development of semi-supervised learning language intelligence technology and Korean tutoring service for foreigners), and supported by the “Advanced GPU Utilization Support Program” funded by the Government of the Republic of Korea (Ministry of Science and ICT).
\bibliographystyle{splncs04}
\bibliography{main}
\end{document}